\documentclass[letterpaper, 10 pt, conference]{ieeeconf}
\usepackage{times}
\usepackage{multicol}
\usepackage[bookmarks=true]{hyperref}

\usepackage{stfloats}
\usepackage{booktabs}

\usepackage{graphicx}
\usepackage[ruled,vlined]{algorithm2e}
\usepackage{mathtools}
\DeclarePairedDelimiter\norm{\lVert}{\rVert}
\usepackage{float} 
\usepackage{amsmath} 

\DeclareMathOperator*{\argmin}{arg\,min}

\usepackage{bbold}
\usepackage{bbm}

\newenvironment{flushenum}{
\begin{enumerate}
  \setlength{\leftmargin}{0pt}
}{\end{enumerate}}

\IEEEoverridecommandlockouts
\begin{document}

\title{Recognizing Orientation Slip\\in Human Demonstrations}

\author{Michael Hagenow$^{1}$, Bolun Zhang$^{1}$, Bilge Mutlu$^{2}$, Michael Zinn$^{1}$, Michael Gleicher$^{2}$


\thanks{*This work was
supported in part by NSF award 1830242 and the University of Wisconsin-
Madison Office of the Vice Chancellor for Research and Graduate Education
with funding from the Wisconsin Alumni Research Foundation.}
\thanks{$^{1}$Michael Hagenow, Bolun Zhang, and Michael Zinn are with the Department of Mechanical
Engineering, University of Wisconsin--Madison, Madison 53706, USA
        {\tt\small [mhagenow|bzhang65|mzinn]@wisc.edu}}%
\thanks{$^{2}$Bilge Mutlu and Michael Gleicher are with the Department of Computer
Sciences, University of Wisconsin--Madison, Madison 53706, USA
        {\tt\small [bilge|gleicher]@cs.wisc.edu}}}


%

\maketitle
\begin{abstract}
Manipulations of a constrained object often use a non-rigid grasp that allows the object to rotate relative to the end effector. This orientation slip strategy is often present in natural human demonstrations, yet it is generally overlooked in methods to identify constraints from such demonstrations. In this paper, we present a method to model and recognize prehensile orientation slip in human demonstrations of constrained interactions. Using only observations of an end effector, we can detect the type of constraint, parameters of the constraint, and orientation slip properties. Our method uses a novel hierarchical model selection method that is informed by multiple origins of physics-based evidence. A study with eight participants shows that orientation slip occurs in natural demonstrations and confirms that it can be detected by our method.
\end{abstract}
\IEEEpeerreviewmaketitle

\section{Introduction}
When people manipulate an object, they frequently use a non-rigid grasp that allows the object to rotate relative to their hand. Such intentional \emph{orientation slip} is often useful as it allows for simpler and less coordinated motions. Our goal is to use human demonstrations to construct robot actions. As a common manipulation strategy, orientation slip must be appropriately considered when interpreting behavior in human demonstrations. We believe that explicitly recognizing orientation slip represents a stepping stone in the direction of considering human strategy when modeling interaction behavior in learning from demonstration (LfD).

\begin{figure}[t]
\centering
\includegraphics[width=3.30in]{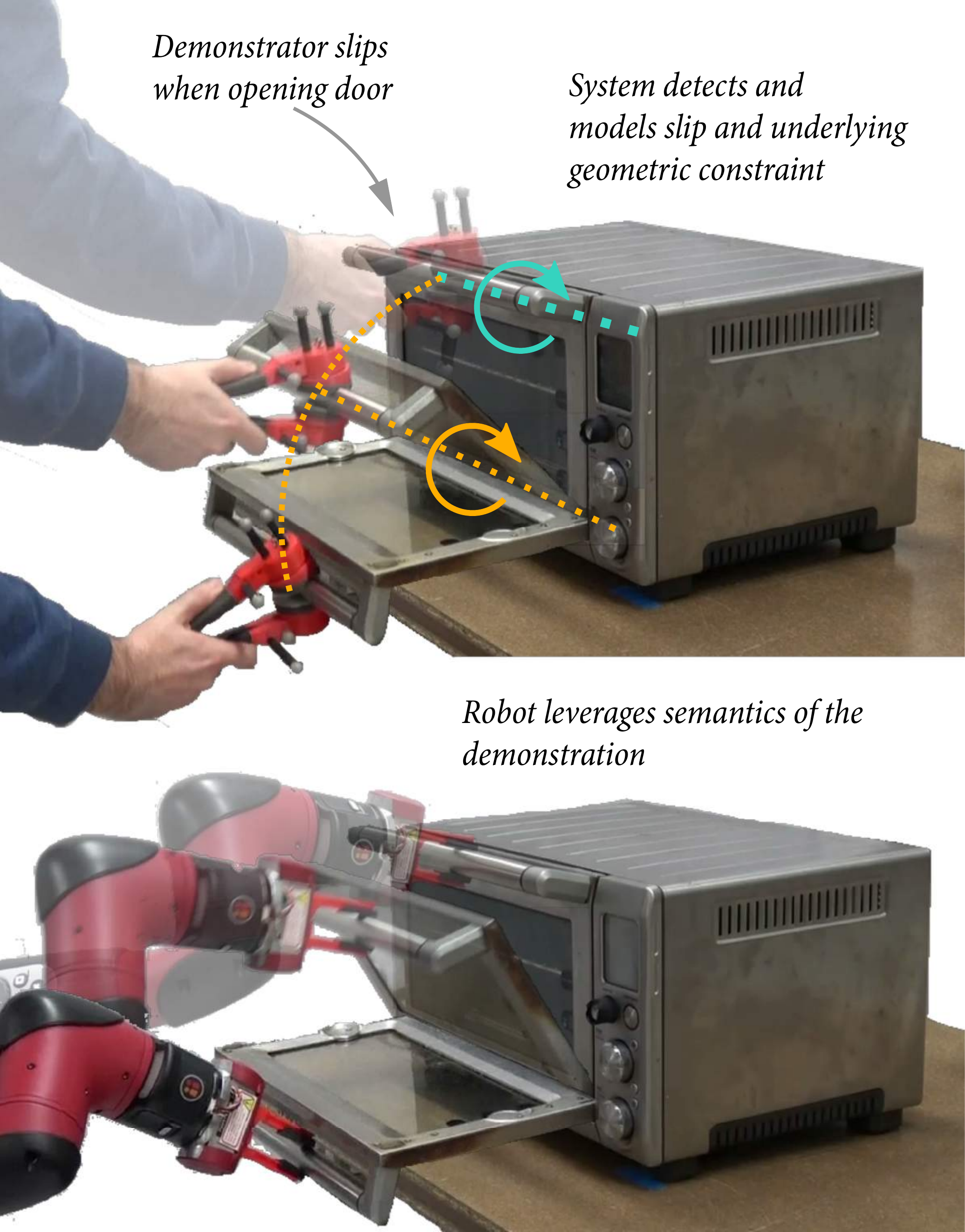}
\caption{Our method recognizes and parameterizes the axis of orientation slip in a human demonstration in addition to the base articulation of the constraint (axial hinge). This semantic understanding of the task interaction can be used to inform a choice from existing robot strategies (e.g., robot controlled slip).}
\label{fig:teaser}
\vspace{-15pt}
\end{figure}

Notably, it is often the case that an orientation slip strategy is used because the grasped object (e.g., handle) is geometrically-constrained to the environment. For example, to open a toaster (Figure \ref{fig:teaser}), the choice to use slip avoids large end-effector motions that would be induced by rigidly grasping the door hinge. Other examples include the prismatic motion of pulling out drawers, the axial motion of turning a crank, and the planar motion of operating an iron. If a robot is to replay or learn from a human demonstration which includes orientation slip, the nature of the constrained interaction must also be identified in addition to the slip properties. Uncovering the semantics of both slip and an underlying geometric constraint can allow a system to determine an appropriate way to execute the task, such as programming the robot to slip or planning a different motion consistent with the permissible articulations of the object.

In this paper, we introduce a method that can model and identify orientation slip in constrained interactions in human demonstrations. Our approach requires only observations of the end effector and can infer the constraint and slip parameters without knowledge of the manipulated object or additional environment sensing. Specifically, our methods take position and wrench trajectories of the end effector as input, for example recorded from instrumented tongs \cite{PraveenaTongs} (Figure \ref{fig:teaser}). Wrench information affords a way to identify relative slip between the end effector and manipulated object via estimates of the friction and grip force, while also providing reaction information that helps to identify underlying constraints. Our method is able to determine the constraint type, parameters, and slip properties using nonlinear fitting coupled with hierarchical model selection. As human demonstrations contain high levels of variability, we conduct a user study to identify the prevalence of orientation slip when users interact with different constraints and to assess the ability of our method to recognize slip interactions.
\section{Related Works}
 The robotics literature has previously demonstrated the value of active slip in robotic manipulation, but has not explored recognizing slip from human demonstrations. Intentional controlled slip can be used to enhance the dexterity of a robot end effector \cite{Brock1988}, reconfigure objects \cite{Gupta1999}, avoid regrasping \cite{Sastry1992}, and improve manipulator kinematics \cite{Sastry1989}. Martin-Martin et al. \cite{martinmartin19} describe a framework for estimating and leveraging likely hand-object articulation using exploratory motion, however the focus is on robot exploration as opposed to learning from human demonstrations.

Orientation slip is a frequent strategy in constrained motion, yet previous methods in learning constrained motion from demonstration and estimating articulation models from interactive perception \cite{bohg2017} have not focused on recognizing and parameterizing orientation slip. In many methods \cite{Joris2014}\cite{Niekum2019}\cite{Joris2008}\cite{ShahClearn}\cite{Niekum2015}\cite{SubramaniRAL}\cite{chou2018}\cite{chou2020}\cite{Armesto2018}, geometric constraints are inferred directly from human demonstrations, but such methods do not explore the semantics of whether slip was present and the resulting impact on constraint models. As an example, Sturm et al. \cite{BurgardSlipHook} study similar interactions with household objects, however, the focus is on rigid grasps, whereas our method can identify both rigid and slip interactions. Moreover, slip during demonstrations violate rigid assumptions present in many constraints and may cause methods to mischaracterize the constraints.

In this work, we aim to uncover the semantics of orientation slip in human demonstrations. We choose to focus on three common constraint models: \emph{axial}, \emph{prismatic}, and \emph{planar}, while speculating our method can be extended to other constraints that occur during manipulation \cite{MorrowConstraints}\cite{Celaya2008}\cite{Knoll2016}. While a more general constraint model, such as encoding the constrained space on a manifold \cite{Berenson2009}\cite{Li2017}, might be able to represent orientation slip, the lack of semantic slip recognition might limit available strategies (e.g., robot controlled slip, replanning) during robot execution. In this work, we leverage the constraint modeling and evaluation approach presented in Subramani et al. \cite{subramani2020method}. However, this work focuses on identifying orientation slip in demonstrations which expands the previous formulation to include new constraint slip models, hierarchical model selection, and an additional model-selection heuristic based on kinetic-friction estimates. 

\section{Modeling and Slip Parameterization}
\label{sec:modeling}

Classifying orientation slip in human demonstrations requires a method to model rotation from orientation slip. In this work, we choose to model the combination of slip and an underlying geometric constraint as a single scleronomic constraint model using a flexible formulation from the multi-body dynamics literature \cite{haug1989computer} which trivializes adding additional constraints (e.g., defining a slip axis) to existing constraint models. Specifically, a constraint consists of equations:
\begin{enumerate}
  \item Constraining the motion of the grasped object to a geometric primitive, referred to as the base articulation.
  \item Defining an additional rotational axis between the grasped object and the end effector provided by the orientation slip of the interaction.
\end{enumerate}

\begin{figure}[t]
\centering
\includegraphics[width=3.1in]{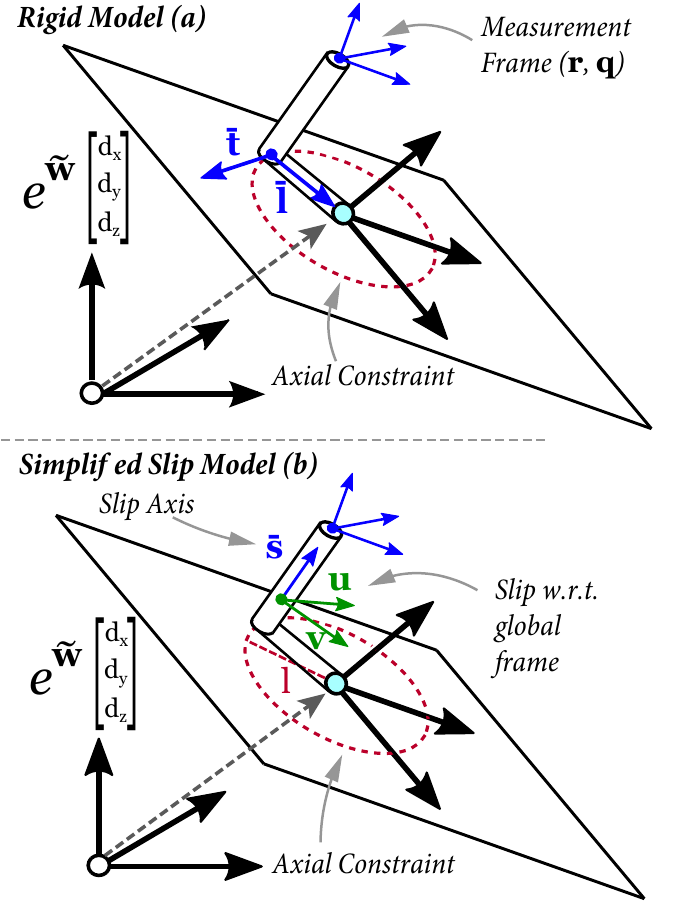}
\caption{The rigid and slip axial models considered in this paper. (a) Rigid model represents a rigid grasp by the measured end effector. (b) Slip model than can be estimated from a single measured pose}
\vspace{-15pt}
\label{fig:axialmodel}
\end{figure}

During orientation slip, the measured pose of the end effector (e.g., hand or instrumented tongs) rotates with respect to the grasped object. As a result, the full kinematics of the constrained interaction consist of the end-effector pose, the grasped object pose, and a global frame of reference. Precisely estimating the additional pose of the grasped object, either through vision-based object detection or additional explicit body tracking, can present practical challenges. We desire to infer the entire interaction purely from measurements of the end effector and impose the restriction that the orientation slip is modeled as a rotational degree of freedom in the global frame, which circumvents the need for a second pose. We argue that this simplified model can still cover many important cases. For example, the handle of a crank (e.g., door window or valve) is often engineered to align with the principle axis of rotation which is static with respect to the environment.

In addition to identifying slip interactions, we desire to be able to differentiate from other types of interactions with the same base articulation (i.e., grasped object). For each base articulation, we consider three interaction models.
\begin{itemize}
  \item \emph{Rigid Model}: The end effector does not slip with respect to the grasped object. This is equivalent to firmly grasping and rotating with the object.
  \item \emph{Slip Model}: The end effector rotates around one axis with respect the grasped object.
  \item \emph{Free Orientation}: The end effector can rotate to any orientation with respect to the grasped object (i.e., ball and socket joint attached to the object).
\end{itemize}
For the base articulation, we consider three common constraints (\emph{planar}, \emph{prismatic}, and \emph{axial}). The approach can be extended to other scleronomic constraint models using a similar constraint description. The formulation of the \emph{axial} models are provided below as an example. Other constraint models were derived following the same procedure starting from the models described in \cite{subramani2020method}.

\subsection{Axial Rigid Model}
An axial motion is when an object rotates around a fixed axis such as opening the door of a refrigerator. In a rigid axial interaction, the object is firmly grasped preventing rotation between the end effector and the object. Thus, the combination of the end effector and grasped object are modeled as a single rigid body that rotates about an axis at a fixed radial distance. The formulation used in this work requires the orientation and location of a plane to define the arc. The plane rotation is represented using two exponential coordinates, $\textbf{w}={\left[ w_{x}\ \ w_{y}\ \ 0 \right]}$ to construct a rotation matrix via the exponential map, $e^{\tilde{\textbf{w}}}\in{SO(3)}$, where $\tilde{\textbf{w}}$ represents the cross product matrix for vector $\textbf{w}$. The radial distance of the arc is defined by a local vector connecting the rigid body to the center of rotation, $\bar{\textbf{l}}$ (Equation \ref{eq:axial1}). This vector and a second orthonormal vector, $\bar{\textbf{t}}$, are restricted to be perpendicular to the normal of the plane, which enforces the orientation of the rigid body (Equations \ref{eq:axial2}-\ref{eq:axial4}). The rigid body is constrained as shown in Figure \ref{fig:axialmodel}(a):
\begin{gather}
{\mathbf{r}+\textbf{A}(\mathbf{q})\bar{\mathbf{l}}-{{[{{d}_{x}},{{d}_{y}},{{d}_{z}}]}^{T}}}=\textbf{0}_3\label{eq:axial1}\\
{{\left( \textbf{A}(\mathbf{q})\bar{\mathbf{l}}~ \right)}^{T}}{{e}^{{\tilde{\mathbf{w}}}}}{{\left[ 0\ \ 0\ \ 1 \right]}^{T}}=0 \label{eq:axial2}\\
{{\left( \textbf{A}(\mathbf{q})\bar{\mathbf{t}}~ \right)}^{T}}{{e}^{{\tilde{\mathbf{w}}}}}{{\left[ 0\ \ 0\ \ 1 \right]}^{T}}=0 \label{eq:axial3}\\
{{\bar{\mathbf{l}}}^{T}}\bar{\mathbf{t}}=0\label{eq:axial4}
\end{gather}
 where $\textbf{r}$ is the measured position,  ${\left[ d_{x}\ \ d_{y}\ \ d_{z} \right]}$ is the center of the arc, and $\textbf{A(q)}$ represents the rotation matrix for the measured orientation (represented by quaternion, $\textbf{q}$). Equations \ref{eq:axial1}-\ref{eq:axial4} define the rigid axial model.

\subsection{Axial Slip Model}
\subsubsection{Base Articulation Model}
The base articulation model enforces that a measured point on the grasped object rotates with a fixed radius around a center point. This is equivalent to the free orientation model since only the grasping point is enforced, which does not restrict orientation of the end effector. A point $\mathbf{r}$ that is consistent with an axial motion satisfies:
\begin{equation}
{{\left( {{\left[ d_{x}\ \ d_{y}\ \ d_{z} \right]}^{T}}-\mathbf{r} \right)}^{T}}{{e}^{{\tilde{\mathbf{w}}}}}{{\left[ 0\ \ 0\ \ 1 \right]}^{T}}=0
\label{eq:axial5}
\end{equation}
\begin{equation}
\norm{{{\left[ d_{x}\ \ d_{y}\ \ d_{z} \right]}^{T}}-\mathbf{r}}_{2}-l^{2}=0
\label{eq:axial6}
\end{equation}
where $l$ is the length of the radial arc. Equations \ref{eq:axial5}-\ref{eq:axial6} define the base articulation (i.e., free orientation axial model). Additional constraints are required to further restrict the motion to an orientation slip interaction.

\subsubsection{Slip Model}
 When the axis of slip can be expressed as static with respect to the environment, such as in the case of the handle of a toaster oven door, the simplified slip formulation can be used (Figure \ref{fig:axialmodel}(b)). In this case, the two vectors, $\textbf{u}$ and $\textbf{v}$, are expressed globally and mutually orthogonal to the slip axis. This allows the slip axis to be fit independently of the base articulation. The slip axis on the rigid body is represented by a local vector, $\bar{\textbf{s}}$:

\begin{gather}
{{\mathbf{u}}^{\textbf{T}}(\mathbf{A(q)}\bar{\mathbf{s}})}=0\label{eq:axial7}\\
{{\mathbf{v}}^{\textbf{T}}(\mathbf{A(q)}\bar{\mathbf{s}})}=0\label{eq:axial8}\\
{{\bar{\mathbf{u}}}^{T}}\bar{\mathbf{v}}=0\label{eq:axial9}
\end{gather}
The combination of the the base articulation model, Equations \ref{eq:axial5}-\ref{eq:axial6}, and Equations \ref{eq:axial7}-\ref{eq:axial9} define the slip axial model.

\section{Recognizing Slip in Demonstrations}
\label{sec:modelselection}

In order to detect orientation slip out of the larger class of physical interactions proposed in Section \ref{sec:modeling}, we desire to determine whether an identified section of a demonstration is rigid, slipping along a single axis, or may be a free orientation granted through a hinge or bearing. We frame this as a problem of model selection, where we desire to find the most likely interaction out of a library of candidate constraint models. By looking at pose and wrench information provided in the demonstration, we construct multiple forms of physics-based evidence that guide selection. The overall method for model selection consists of four steps:
\begin{flushenum}
  \item The three base articulation models are fit and evaluated to determine the most likely base articulation. 
  \item For the selected base articulation, the three orientation articulations are evaluated based on orientation and moment residuals to construct model probabilities.
  \item A second probability for the three orientation articulations is constructed based on kinetic friction estimates.
  \item The final model is selected using the base articulation model and combining the probabilities for each orientation model.
\end{flushenum}

\subsection{Hierarchical Model Selection}
Since many of our models use the same base articulation with different orientation interactions, there is precedence for first identifying the most likely base articulation of a constraint. Once the base articulation is selected, a second step can determine the nature of admissible rotation (i.e., the interaction model), including potential orientation slip.

First, all potential models are fit to the recorded data, $\mathbf{\mathcal{D}}$, using the iterative reweighted least squares (IRLS) optimization procedure detailed in \cite{subramani2020method}. We provide a partial review of the procedure to contextual the hierarchical model selection method. Since some interactions (e.g., rigid orientation) can generate reaction moments, only the kinematic relationships defined by the constraint equations are used in the fitting. The model parameters are fit according to: 
\begin{equation}\label{eq:kinematicfitting}
    \hat{\boldsymbol{\alpha}}_{MLE} = \argmin_{\boldsymbol{\alpha}}(\sum\limits_{i = 1}^{N}{w_i\|\Phi(\mathbf{p}_i,\boldsymbol{\alpha})\|})
\end{equation}  
where $\boldsymbol{\alpha}$ are the constraint parameters, $\mathbf{\Phi}$ is a row vector of constraint equations for the constraint model, $\textbf{p}_{i}$ is the measured pose (transformed to the grasp point), and $w_{i}$ is the weight for sample $i$ that is adjusted during the re-weighting process of the robust fitting.

Because the constraint equation residuals for each model do not have consistent units, the base articulation model is selected by computing separate fitting errors based on the least squares results. A position error is constructed by running a separate optimization (described in \cite{subramani2020method} to find the pose closest to the measured pose that is also consistent with the constraint parameters (Note: estimates of the closest consistent position and orientation, $\hat{\textbf{r}}^{*}$ and $\hat{\textbf{q}}^{*}$ respectively, are solved simultaneously). In addition to position error, a force residual based on permissible reaction forces is constructed via a reduced form of the rigid body equations of motion where inertia is considered negligible.
\begin{gather}
E_{\mathbf{r}} =  ||\hat{\mathbf{r}}^\star - \mathbf{r}||\\
E_{\mathbf{f}} =   ||\boldsymbol{\Phi} _{\mathbf{r}}^{T}\boldsymbol{\lambda}-\textbf{f}_{\mu}+\mathbf{f}|| / ||\mathbf{f}|| 
\end{gather}
where $\boldsymbol{\Phi}_{\mathbf{r}}$ is the constraint Jacobian, $\boldsymbol{\lambda}$ are the Lagrange multipliers, $\hat{\textbf{r}}^{*}$ is the closest position consistent with the constraint, $\textbf{r}$ is the measured position, $\textbf{f}$ is the measured force, and $\textbf{f}_{\mu}$ is the friction force which is conservatively estimated as all forces in the direction of motion (see \cite{subramani2020method}). The force residual is equivalent to the percentage of forces inconsistent with the constraint model. While orientation and moment residuals are available via a similar method, only position and force residuals are considered in the first step of selection since the orientation has not yet been considered.

The base articulation models are compared based on the percentage of position and force residual samples that are above the estimated error of the system (e.g., motion capture resolution, force resolution), which are defined below with subscript \emph{SE}. While an energy-based method can also combine residuals with different units, such a method can still have practical issues that require weighting based on the relative error levels of pose and wrench measurements. The most likely base model is determined as:
\begin{equation}
    m = \argmin_{m\in \mathcal{M}}(\sum_{i}(\mathbb{1}_{E_r(i,m)>r_{SE}}+\mathbb{1}_{E_f(i,m)>f_{SE}})
\end{equation}

Once the base articulation model is selected, the second step calculates a probability for each interaction model via the orientation and moment residuals from the fitting process. The orientation metric is constructed as the angle between the measured orientation and the closest permissible orientation consistent with the constraint, $\textbf{q}^{*}$. The moment error is defined via reduced equations of motion using the same Lagrange multipliers, $\mathbf{\boldsymbol{\lambda}}$, as above.
\begin{gather}
E_{\mathbf{q}} =  \angle\left(conj(\mathbf{q})\star \hat{\mathbf{q}}^\star\right)\\
E_{\mathbf{n}} =   ||\mathbf{\Phi} _{\boldsymbol{\pi} }^{T}\boldsymbol{\lambda}-\textbf{n}_{\mu}+\mathbf{n}||
\end{gather}
where $\angle$ is the angle between rotations computed through quaternion multiplication ($\star$),  $\textbf{n}$ is the measured moment, $\textbf{n}_{\mu}$ is the friction moment (see \cite{subramani2020method}), and $\mathbf{\Phi}_{\boldsymbol{\pi}}$ is the constraint rotation Jacobian \cite{haug1989computer}.

 Since the free orientation model does not allow reaction moments (which without friction could lead to zero measured moments), the moment error is not normalized and instead expressed in newton-meters. The probability for the slip and rigid models is defined based on the percentage of residual values above the estimated measurement error (combination of DAQ and mechanical tolerances) of the system. The posterior probability for model, $m$, can be expressed via Bayes Theorem. In this case, there is no prior (i.e., the models are equally likely), but the expression is left in Bayesian form to be consistent with \S\ref{sec:kinetic}):

\begin{equation}\label{eq:MS}
P(\mathcal{D}|m)=\left(\sum_{i} \left(
\begin{aligned}
       &\mathbb{1}_{D_q(i)>q_{SE}} \\
       &+ \mathbb{1}_{E_n(i)>n_{SE}} \\
\end{aligned}
\right)+1\right)^{-1}
\end{equation}

\begin{equation}
    P( m \mid \mathcal{D} ) =  \frac{P(\mathcal{D}|m)}{\sum_j{P(\mathcal{D}|\mathcal{M}_{j})}}
    \end{equation}

While this probability is capable of differentiating the orientation models, we construct a second posterior probability based on kinetic friction to more robustly recognize rigid, slip, and free interactions.

\subsection{Evidence of Slip from Kinetic Friction}
\label{sec:kinetic}
A Coulomb friction model dictates that during orientation slip, static friction between the end effector and rigid body will break resulting in a friction force that is proportional to the applied normal grip force. Our key idea is to extract an estimate of these forces and use this physical relationship to construct a second slip probability.

\begin{figure}[t]
\centering
\includegraphics[width=3.2in]{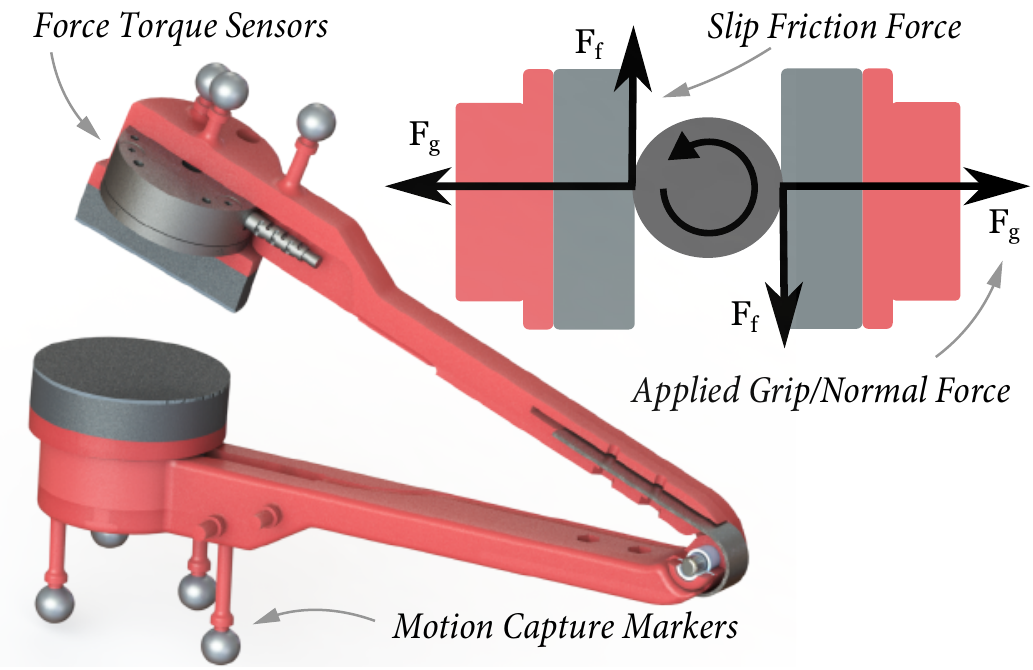}
\vspace{-5pt}
\caption{Using some assumptions about the geometry, friction and grip forces that would normally cancel out can be extracted from the tongs to provide additional evidence of slip.}
\label{fig:tongs}
\vspace{-10pt}
\end{figure}

In our work, we use instrumented tongs where each pad is instrumented with a force-torque sensor. The magnitudes of grip and friction forces can be estimated from measured data by assuming a point interaction with parallel gripping (i.e., the vectors are rotated by a negligible angle, see Figure \ref{fig:tongs}). The friction and grip components are estimated as:

\begin{gather}
F_{c} = \frac{1}{2}\left(
\begin{aligned}
    ||{\textbf{F}_{L}}\cdot{\hat{\textbf{x}}_{Lc}}||+||{\textbf{F}_{R}}\cdot{\hat{\textbf{x}}_{Rc}}||\\
    -||{\textbf{F}_{L}}\cdot{\hat{\textbf{x}}_{Lc}}+{\textbf{F}_{R}}\cdot{\hat{\textbf{x}}_{Rc}}||
\end{aligned}\right), c\in \{f,g\}\label{eq:kinetic1}
\end{gather}
where $L$ and $R$ represent the left and right tong respectively. The $x$ vectors represent unit vectors in the relevant directions ($f$ is friction, $g$ is grip/normal) with respect to the force torque sensor frames. Other input methods with redundant interaction force measurements might be able to provide similar estimates.
By normalizing the friction forces by the normal forces (i.e.,  $F_{f}/F_{g}$), the sliding friction coefficient, $\hat\mu_{k}(i)$, is estimated for each sample $i$. The slip probability is constructed based on intuition of expected forces:
\begin{enumerate}
  \item Evidence of slip will be characterized by a constant value of $\hat\mu_{k}(i)$ equal to the sliding friction coefficient. 
  \item Evidence of a free model should have zero friction and consequently, a zero value of $\hat\mu_{k}(i)$,
  \item A rigid grasp can register internal forces that are indistinguishable from friction. Any value of $\hat\mu_{k}(i)$ less than the static friction coefficient is permissible.
\end{enumerate}

\begin{figure}[]
\centering
\includegraphics[width=3.0in]{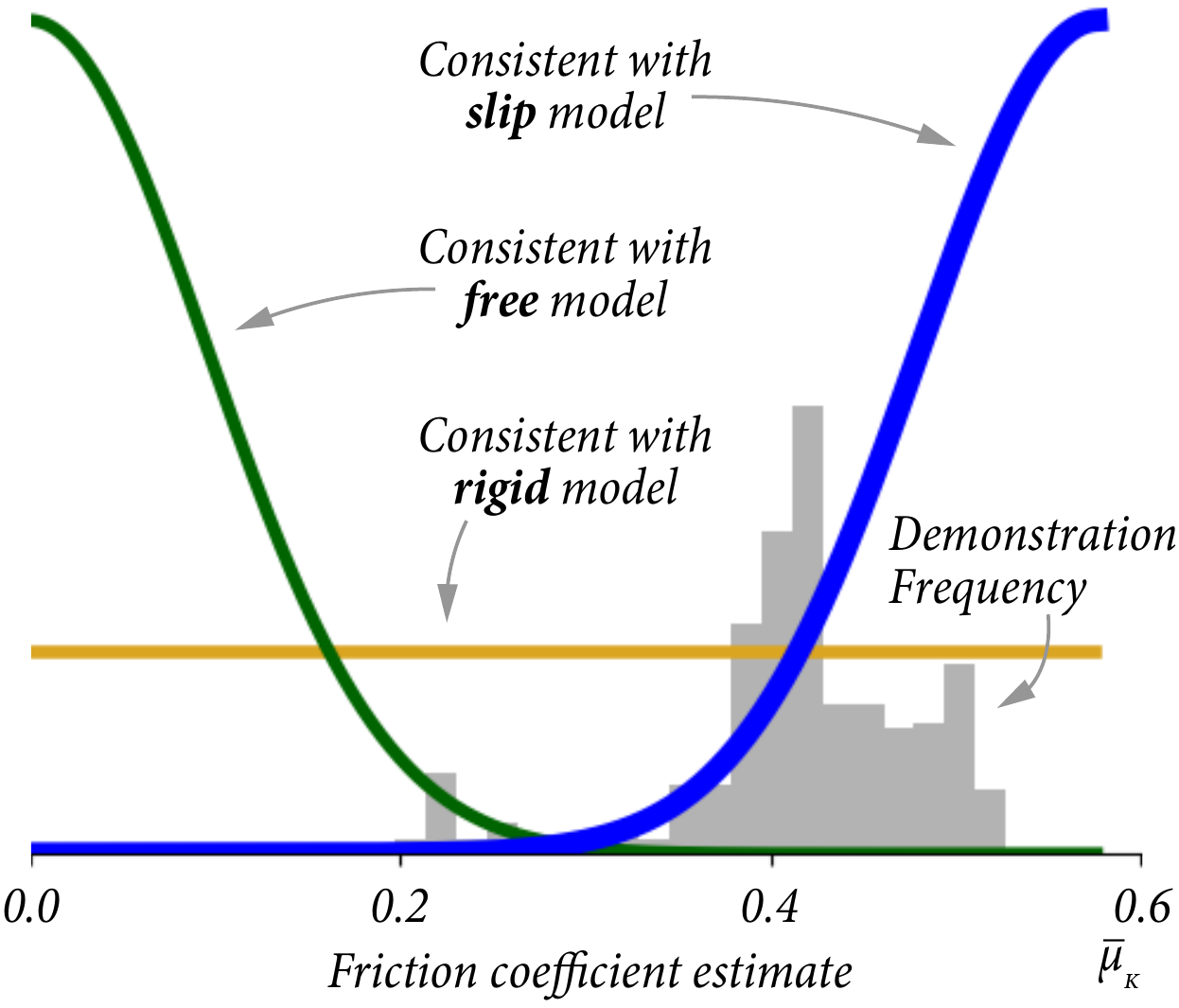}
\caption{Using the friction and normal forces, the value of the sliding friction coefficient is estimated for a slip demonstration. The model probability is constructed by fitting distributions based on expected values for the models. In this case, the frequency best aligns with the slip model Gaussian (blue).}
\label{fig:kineticfriction}
\vspace{-10pt}
\end{figure}

To quantify the above behavior, we use a Gaussian-weighting approach. The key idea is to weigh each $\hat\mu_{k}(i)$ against the expected behavior of each orientation articulation, which can be used to construct a posterior probability. Since the free and slip models are expected to be at the minima and maxima respectively, the expected values are represented with half-normal distributions. Since the rigid model can assume any value, it is represented by a uniform distribution. The expected value of the sliding friction, $\bar\mu_{k}$, is set to 0.58 based on separately-collected empirical measurements. The probability of each model is defined as follows:

\begin{gather}
P(x\mid\hat{\mu}_{k}) = \frac{\left(\sum_{i}(f_{x}(\hat\mu_{k}(i))\right)P(x)}{\sum_{x}\left(\sum_{i}(f_{x}(\hat\mu_{k}))\right)P(x)},x\in\{\emph{slip},\emph{rigid},\emph{free}\}
\end{gather}
\begin{gather}
f_{slip}(\bar\mu_{k}(i))=\mathcal{N}_{x}(\hat\mu_{k}(i),\mu=\bar\mu_{k},\sigma=\bar\mu_{k}/6)\\
f_{free}(\bar\mu_{k}(i))=\mathcal{N}_{x}(\hat\mu_{k}(i),\mu=0,\sigma=\bar\mu_{k}/6)\\
f_{rigid}(\bar\mu_{k}(i))=1.0
\end{gather}
where $P(x)$ is the model prior (which is non-normalized and can be interpreted as a weighting factor) and $\mathcal{N}_{x}$ is the Gaussian function centered at $x$ with mean, $\mu$ and variance, $\sigma$. Figure \ref{fig:kineticfriction} shows $\hat\mu_{k}(i)$ estimates for an example slip demonstration. $P(x)$, can be tuned based on knowledge of the application. In our testing, we found that free orientation demonstrations had $\hat\mu_{k}$ values consistently higher than zero. To avoid these being incorrectly classified as a rigid model, we reduced the prior of the rigid model. Our final values were: $P_{slip}=P_{free}=1,P_{rigid}=0.625$. Equating distribution areas as shown in Figure \ref{fig:kineticfriction} also assumes that all measured values range between 0 and $\bar{\mu_{k}}$, which is not necessarily the case. After the base articulation model is selected, there are multiple forms of evidence to assess the orientation restrictions of the orientation articulation. The final selection uses the product of the two probabilities.
\section{Experimental Evaluation} \label{sec:experiments}
To evaluate the robustness of our method against the variability present in human demonstrations, we performed a user study involving eight participants with no prior knowledge of or prior experience with the research recruited from a university campus. The study followed a within-subjects design where participants completed two sets of demonstration tasks in a counter-balanced order.

 
\begin{figure}[]
\centering
\includegraphics[width=3.3in]{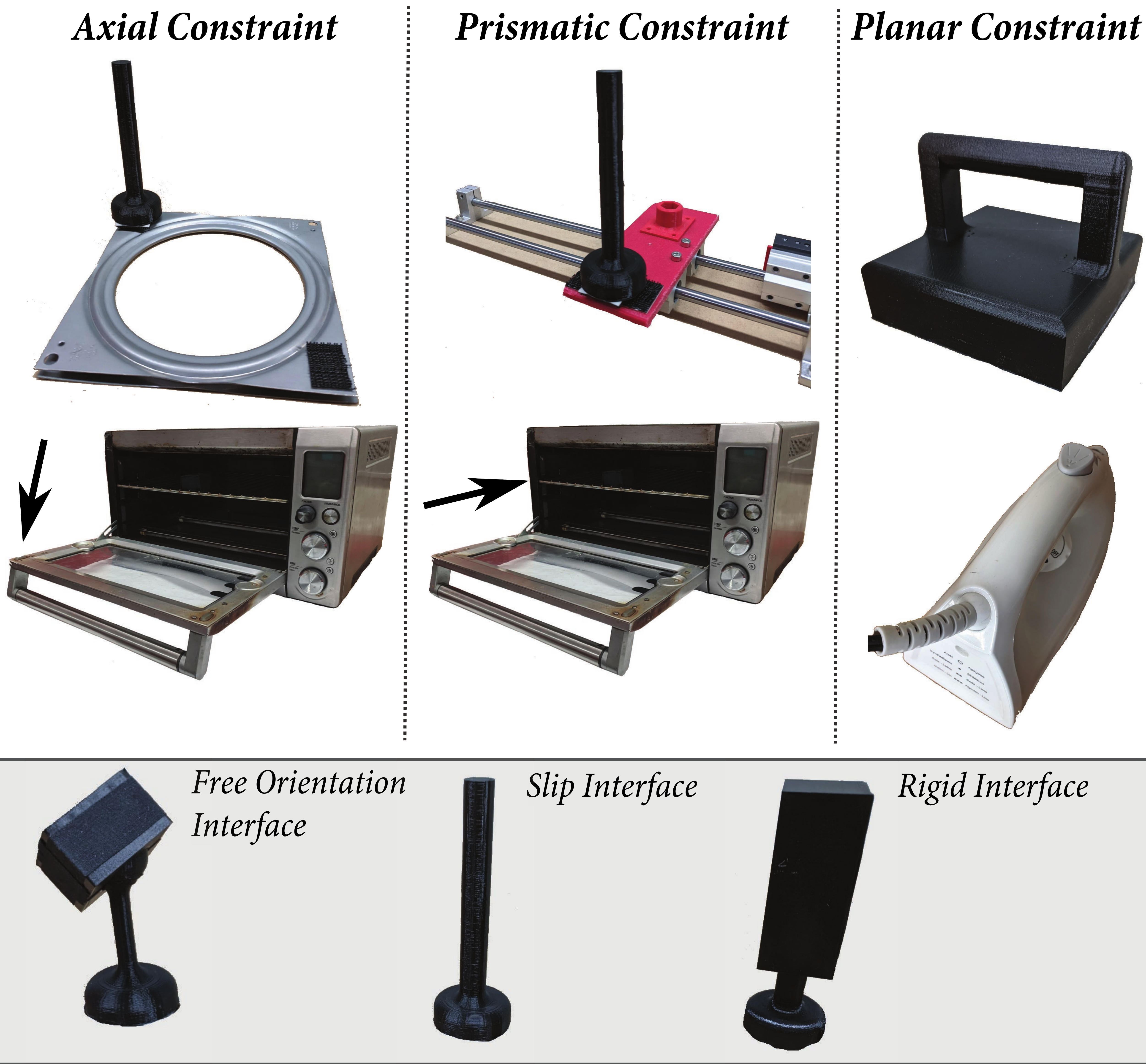}
\vspace{-5pt}
\caption{ Experimental setup of constraints. \textit{Top:} six constraints are used, three of which were primitive and three involved interacting with home appliances (\S\ref{sec:slipprevalence}). \textit{Bottom (gray):} Three interfaces used for interacting with constraints (\S\ref{sec:expmodelselection}).}
\vspace{-10pt}
\label{fig:constraints}
\end{figure}

\subsection{Study Tasks}
\subsubsection{Task 1. Prevalence of Slip in Demonstrations}
\label{sec:slipprevalence}
In the first demonstration task, we asked unprompted participants to engage in six constrained interactions (Figure \ref{fig:constraints}) to establish whether orientation slip occurs in natural demonstrations. These interactions involved manipulating three primitive constraints that were interfaced with a black cylindrical handle and three home appliances, including a toaster oven door, a toaster oven rack, and a household clothing iron. Demonstrations where slip was the predominant manipulation strategy based on observations were labeled as ``slip.''

\subsubsection{Task 2. Identifying Slip in Constrained Interactions}
\label{sec:expmodelselection}
The second demonstration task aimed to show that our hierarchical model selection method can differentiate constraint and interaction models in human demonstrations. Different interfaces, show in Figure \ref{fig:constraints}, were attached to the three primitive constraints. Users were instructed to excite the permissible motion of the interface and constraint. Each participant provided two five-second demonstrations for each of the interfaces connected to each of the three constraint primitives. In order to be consistent with the simplified slip model, the planar slip model had the handle oriented parallel with the plane normal to remain static in the global frame. 

\subsection{Results}
\subsubsection{Task 1. Prevalence of Slip in Demonstrations}
Our data, summarized in Table \ref{table:slipcount}, indicates that orientation slip was common in constraints that induced a large, undesirable variation in orientation. Slip was used in nearly all interactions with axial motion. With the sliding prismatic platform, many participants centered their wrist and used wrist rotation and orientation slip to move the handle back and forth to minimize wrist translation. Because the oven rack prismatic constraint was in an enclosed space and grasped from the end, a slip strategy was infrequently used. Slip was not common in planar interactions. While slip can help extend the reach of motion, we speculate that participants had a tendency to use a rigid grasp to apply sufficient force to keep the object in contact with the table.

\begin{table}[!b]
\centering
\caption{Frequency of slip in unprompted demonstrations.}\label{table:slipcount}
\begin{tabular}{@{}cccccc@{}}
\toprule
Constraint & Hand & Tongs & Constraint & Hand & Tongs \\
\midrule
Axial & 8/8 & 8/8 & Oven Door & 8/8 & 7/8 \\ 
Prismatic & 8/8 & 5/8 & Baking Rack & 0/8 & 0/8 \\ 
Planar & 1/8 & 0/8 & Iron & 0/8 & 0/8 \\ 
\bottomrule
\end{tabular}
\end{table}

Our data provides evidence that an orientation slip strategy is used in interactions with different types of geometric constraints, including everyday objects. We believe that many other day-to-day objects naturally elicit orientation slip.

\subsubsection{Task 2. Identifying Slip in Constrained Interactions}
\begin{figure}[]
\centering
\includegraphics[width=3.2in]{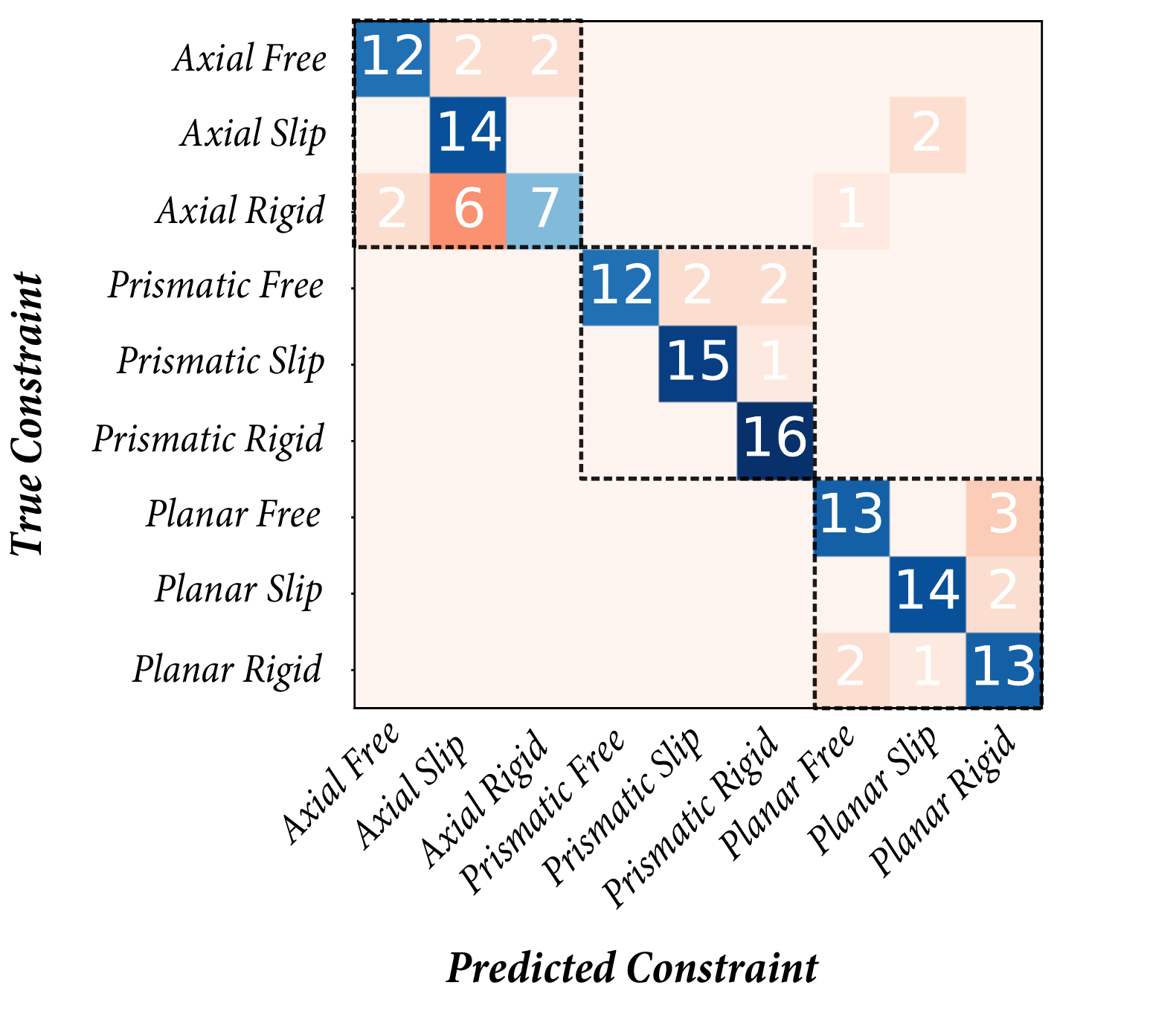}
\vspace{-10pt}
\caption{Our model selection method has a classification accuracy of 0.81.The base model classification accuracy is 0.98 (Black dotted boxes).}
\label{fig:confusionresults}
\vspace{-15pt}
\end{figure}

Data from the second task are summarized in Figure \ref{fig:confusionresults}. Our fitting method was consistently able to fit the constraints. The accuracy was verified by comparing each demonstration to an aggregate ground truth (i.e., all demonstrations with the same base articulation). The median rotation error (e.g., exponential map) was 0.032 rad and the median distance error (e.g., radius, center of arc) was 0.0094m. We observed that the system consistently identified the correct base articulation (i.e., \emph{axial}, \emph{prismatic}, or \emph{planar}) with a classification accuracy of 0.98. The overall classification accuracy of the constraint and interaction type was 0.81. The true-positive recognition rate for orientation slip interactions was 94\%.

We found nearly half of rigid axial interactions to be incorrectly classified as slip axial interactions. However, due to the difficulty of performing an axial motion rigidly, participants often showed small amounts of unavoidable orientation slip. Both the hierarchical selection and combination of kinetic friction evidence led to increases in classification accuracy. Using the four error residuals alone resulted in a classification accuracy of 0.74. Simply evaluating the residuals in a hierarchical manner by selecting the base articulation first increased the accuracy to 0.76. The hierarchical evaluation also increased the base model accuracy from 0.94 to 0.98.

Our findings show the value of hierarchical and redundant model selection. We also show that orientation slip can be identified in most demonstrations by minimally-trained users.


\section{General Discussion}
In this paper, we presented a method for identifying prehensile orientation slip in constrained interactions. However, there were a number of limitations that serve as inspiration for future work. 
The method assumes a consistent object grasp point throughout the articulation. It is also assumed that the slip direction can be expressed as a static direction in the global frame, which would not work for tasks like ironing, where the slip direction rotates with the articulated object (i.e., iron). Finally, the kinetic friction estimate assumes parallel gripping with redundant force torque information. In future work, we plan to consider multi-pose estimation (i.e., non-static slip direction), more advanced friction and constraint models, and additional input devices.


\bibliographystyle{IEEEtran}
\bibliography{references}

\end{document}